%% file: pds-classifier-21.tex
\def\year{2021}\relax
\documentclass[letterpaper]{article} 
\usepackage{aaai21}  
\usepackage{times}  
\usepackage{helvet} 
\usepackage{courier}  
\usepackage[hyphens]{url}  
\usepackage{graphicx} 
\usepackage{natbib}  
\usepackage{caption} 
\usepackage[externallead,year=2021]{jplack}
\usepackage[group-separator={,},group-minimum-digits={3}]{siunitx}
\usepackage{acronym}
\usepackage{booktabs}
\usepackage{subfigure}
\usepackage [english]{babel}
\usepackage [autostyle, english = american]{csquotes}
\usepackage{amsmath}
\MakeOuterQuote{"}
\title{Mars Image Content Classification: \\
Three Years of NASA Deployment and Recent Advances}
\author{
    Kiri Wagstaff\textsuperscript{\rm 1},
    Steven Lu\textsuperscript{\rm 1},
    Emily Dunkel\textsuperscript{\rm 1},
    Kevin Grimes\textsuperscript{\rm 1},
    Brandon Zhao\textsuperscript{\rm 2}, 
    Jesse Cai\textsuperscript{\rm 3}, \\
    Shoshanna B. Cole\textsuperscript{\rm 4},
    Gary Doran\textsuperscript{\rm 1},
    Raymond Francis\textsuperscript{\rm 1},
    Jake Lee\textsuperscript{\rm 1}, and
    Lukas Mandrake\textsuperscript{\rm 1}
    \\
}
\affiliations{
    \textsuperscript{\rm 1} Jet Propulsion Laboratory, California
    Institute of Technology, 4800 Oak Grove Drive, Pasadena, CA,
    91109-8099, USA \\ 
    \textsuperscript{\rm 2} Duke University, 2138 Campus Drive,
    Durham, NC 27708, USA \\ 
    \textsuperscript{\rm 3} California Institute of Technology, 1200 E California Blvd, Pasadena, CA 91125, USA \\
    \textsuperscript{\rm 4} Space Science Institute, 4765 Walnut St, Suite B, Boulder, CO 80301, USA \\
    \{kiri.l.wagstaff,you.lu,emily.dunkel,kevin.m.grimes\}@jpl.nasa.gov, 
    brandon.zhao@duke.edu,
    jycai@caltech.edu,
    scole@spacescience.org,
    \{gary.b.doran.jr,raymond.francis,jake.h.lee,lukas.mandrake\}@jpl.nasa.gov
}

\usepackage{color}
\newcommand{\todo}[1]{{\textcolor{blue}{{[#1]}}}}
\newcommand{\comment}[1]{}

\begin{document}

\acrodef{MSL}{Mars Science Laboratory}
\acrodef{MER}{Mars Exploration Rovers}
\acrodef{HiRISE}{High Resolution Imaging Experiment}
\acrodef{Mastcam}{Mast Camera}
\acrodef{MAHLI}{Mars Hand Lens Imager}
\acrodef{Pancam}{Panoramic Camera}
\acrodef{CNN}{Convolutional Neural Network}
\acrodef{PDS}{Planetary Data System}
\acrodef{Imaging Node}{Cartography and Imaging Sciences Node}
\acrodef{EDR}{Experiment Data Record}
\acrodef{RDR}{Reduced Data Record}
\acrodef{IDAR}{Interactive Data Analyzer and Reviewer}
\acrodef{MRO}{Mars Reconnaissance Orbiter}
\acrodef{v1}{version 1}
\acrodef{v2}{version 2}

\maketitle

\begin{abstract}
The NASA Planetary Data System hosts millions of images acquired from
the planet Mars. To help users quickly find images of
interest, we have developed and deployed content-based
classification and search capabilities for Mars orbital and surface
images. The deployed systems are publicly
accessible using the PDS Image Atlas. We describe
the process of training, evaluating, calibrating,
and deploying updates to two CNN classifiers for images collected by
Mars missions. 
We also report on three years of deployment
including usage statistics, lessons learned, and plans for the future.
\end{abstract}

\input{pc21-introduction}
\input{pc21-related-work}
\input{pc21-dataset}
\input{pc21-classifier}
\input{pc21-calibration}
\input{pc21-evaluation}
\input{pc21-deployment}
\input{pc21-conclusion-future-work}

\section{Acknowledgments}
We thank Michael McAuley from the \ac{PDS} \ac{Imaging Node} for the
continuing support of this work and Anil Natha for assistance with the
Google Analytics results.
We also thank the numerous volunteers who helped label the Mars images. 
\jplack{}
This publication uses data generated via the Zooniverse.org platform,
development of which is funded by generous support, including a Global
Impact Award from Google, and by a grant from the Alfred P. Sloan
Foundation.  

\bibliography{refs}

\end{document}

%% file: pc21-introduction.tex
\section{Introduction}

The NASA \ac{PDS} maintains archives of data
collected by NASA missions that explore our solar system.  The
\ac{PDS} \ac{Imaging Node} provides access to millions of
images of planets, moons, comets, and other bodies.  Given the large
and continually growing volume of data, there is a need for tools that
enable users to quickly search for images of interest.  Each image
product is described by a rich set of searchable metadata properties
such as the time it was collected, the instrument used, the image target,
local season, 
etc.  

However, users often wish to search on the {\em content} of the image
to zero in on those images most relevant to a scientific investigation
or individual curiosity.  Manually searching through millions of
images is infeasible.  In previous work, we trained image classifiers
to detect classes of interest in Mars orbital and
surface images~\cite{wagstaff:deepmars18}.  Using the predictions made
by these classifiers, 
users can interactively search for classes of interest using the PDS Image
Atlas\footnote{\url{https://pds-imaging.jpl.nasa.gov/search/}}.
Since the deployment of these classifiers in late 2016 and through
August 2020, their
predictions have been used to satisfy \num{62613} searches 
on the Atlas website.  

In this paper, we report on several new advances within this domain.
First, we expanded the set of classes known to each classifier to
broaden their coverage of different content types.  Second, we
employed classifier calibration to produce more reliable posterior
probabilities, which is vital since only classifications with a
posterior probability of at least 0.9 are displayed to users.  
\comment{
Third,
we created a new classifier for the Mars Exploration Rover image
archive, which provides the first opportunity to search this massive
data set by content.
}
Finally, we now report on three years of
deployment including usage statistics, lessons learned, and plans for
the future.


%% file: pc21-related-work.tex
\section{Related Work}


Machine learning image classification has achieved high levels of
performance since the adoption of convolutional neural networks (CNNs)
trained on millions of images~\cite{krizhevsky:alexnet12}.  In
addition to demonstrated improvements in accuracy, the use of a CNN
removes the need for manual feature engineering.  The ability to adapt
or ``fine-tune'' large networks enables the re-use of learned lower
levels of the network on new image collections while customizing the
output nodes to the classes of interest.  \citet{palafox:cnn17}
showed that a CNN out-performed a support vector machine classifier on
finding Mars landforms of interest.
In previous work, we demonstrated the ability to fine-tune the AlexNet
classifier for application to images collected by instruments in Mars
orbit and on the Mars surface~\cite{wagstaff:deepmars18}.
Other approaches with relevance for planetary exploration are terrain
classification of regions within an image to inform
navigation~\cite{rothrock:spoc16} 
and generating text captions for planetary images and enable a larger
search vocabulary~\cite{qiu:scoti-caption20}, as opposed to a fixed
set of image classes.

%% file: pc21-dataset.tex
\section{New Mars Classifier Data Sets}

We created two new labeled data sets to train and evaluate the latest
versions of our Mars image classifiers.
The HiRISE 
images were collected by the \ac{HiRISE} instrument
onboard the \ac{MRO}~\cite{mcewen:hirise07}, while the MSL 
images were collected by the \ac{Mastcam}
and \ac{MAHLI} instruments mounted on \ac{MSL} Curiosity
rover~\cite{grotzinger:mslmission2012}. To ensure high
quality, the labels for both data sets were acquired using crowdsourcing
with local volunteers who received specific training for each data
set. 

\input{pc21-hirise-dataset.tex}
\input{pc21-msl-dataset.tex}

%% file: pc21-hirise-dataset.tex
\subsection{HiRISE Orbital Data Set (v3)}

In previous work~\cite{wagstaff:deepmars18},
we compiled \num{3820} images of Mars surface features that covered five
classes of interest.  The new HiRISE data set (v3)
increases the number of labeled images to \num{10815} (before
augmentation), with eight
classes~\cite{dataset:hirise20}\footnote{\url{https://doi.org/10.5281/zenodo.4002935}}. 

\comment{
The HiRISE image data set was created by cropping out portions of
large images taken in long strips as HiRISE sweeps over the planet,
with the primary goal of studying martian surface processes and
landscape evolution. Overall image sizes have a width of $6$ km by a
programmable length of up to $60$ km, with a resolution of $30$
centimeters/pixel. For more information, please see
\url{https://mars.nasa.gov/mro/mission/instruments/hirise/}.
}

HiRISE images consist of long strips that cover up to \num{60} km with
a \num{6}-km wide swath at a resolution of $30$ centimeters/pixel.  
To identify surface features of interest, we employed a focus of
attention mechanism known as {\em dynamic
  landmarking}~\cite{wagstaff:landmarks12}.  This process scans
through a large image to identify visually salient regions, which are
termed {\em landmarks}.  The salience of each pixel is defined as a
linear combination of the response to a Canny filter and the
Earth mover's distance~\cite{rubner:emd98} between the distribution of
pixel intensity 
values within a window around the pixel compared to values within a
larger enclosing window.  We employed a genetic algorithm to optimize
the parameters (analysis window sizes, weighting of individual
filters, and salience threshold) based on fourteen HiRISE images with
hand-labeled salient regions.  The salient landmarks within an image
were obtained by identifying connected components of regions that
exceed the salience threshold.
We cropped the salient landmarks from the ``browse'' (reduced
resolution) version of each HiRISE image using a square 
bounding box plus a \num{30}-pixel border, then resized each image to
\num{227} $\times$ \num{227} pixels. 

\begin{table}
\begin{center}
\begin{tabular}{l|ccc}
\bf Class Name & \bf Count & \bf Percent \\ \hline
Bright dune  & 250 & 2.31\%   \\
Crater  & 794 & 7.34\%   \\
Dark dune  & 166 & 1.53\%   \\
Impact ejecta  & 74 & 0.68\%   \\ 
Other  & 8,802 & 81.39\%   \\
Slope streak  & 267 & 2.47\%   \\
Spider  & 164 & 1.52\%   \\
Swiss cheese  & 298 & 2.76\%   \\ \hline
Total & 10,815 & 100\% \\ \hline
\end{tabular}
\end{center}
\caption{HiRISE (Mars orbital) data set class distribution.}
\label{tab:hirise-dist}
\end{table}

The resulting HiRISE image data set contains \num{10815} landmark
images derived
from \num{232} separate HiRISE source images. The class distribution is shown
in Table~\ref{tab:hirise-dist} in alphabetical order. The classes
are highly imbalanced, with the majority of images classified as ``Other''
\comment{(81\%)} and ``Impact ejecta'' \comment{(1\%)} the least
common class.  

\comment{ To get even the small number of "Impact ejecta" in the data set, we
had performed an additional targeted search, sifting through 41 HiRISE source
images from years 2017-2020 that were known to contain this class. After running
our salience detector on these large images, we obtained 186 landmark images, of
which only 26 were found to be "Impact ejecta". This due in part to this class
being found less frequently by our salience detector. The data collection
process is by far the most time intensive part of the process, and we plan to
focus on this and other minority classes for future collects.  }

\begin{figure}
\begin{center}
\subfigure[\scriptsize Bright dune]
{\includegraphics[height=0.78in]{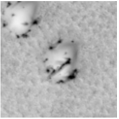}}
\subfigure[\scriptsize Crater]
{\includegraphics[height=0.78in]{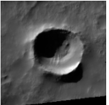}}
\subfigure[\scriptsize Dark dune]
{\includegraphics[height=0.78in]{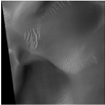}}
\subfigure[\scriptsize Impact ejecta]
{\includegraphics[height=0.78in]{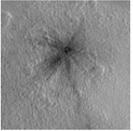}}
\subfigure[\scriptsize Other]
{\includegraphics[height=0.78in]{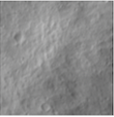}}
\subfigure[\scriptsize Slope streak]
{\includegraphics[height=0.78in]{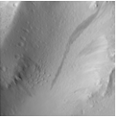}}
\subfigure[\scriptsize Spider]
{\includegraphics[height=0.78in]{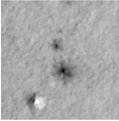}}
\subfigure[\scriptsize Swiss cheese]
{\includegraphics[height=0.78in]{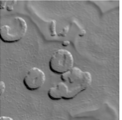}}
\caption{Examples of each class in the HiRISE v3 data set.}
\label{fig:hirise-data}
\end{center}
\end{figure}

\comment{ Our image classes are shown in Figure~\ref{fig:hirise-data}. For our
current work, we combine "spider" and "fan" into a single "spider" class. We
label them separately for future work where we may consider them as separate
classes, due to their distinct visual features. In total, we have eight classes.
}

Examples of each class are shown in Figure~\ref{fig:hirise-data}. 
``Bright dune'', ``Crater'', ``Dark dune'', ``Other'', and ``Slope streak''
classes were included in the v1 HiRISE data set.
``Bright dune'' and ``Dark dune'' are two sand dune classes found on
Mars.  Dark dunes are completely defrosted, whereas bright dunes are
not. 
%
The ``Crater'' class consists of crater images in which the diameter of the
crater is greater than or equal to 1/5 the width of the image and the circular
rim is visible for at least half the crater's circumference.
The ``Slope streak'' class consists of images of dark flow-like features on
slopes. These features are believed to be formed by a dry process in which
overlying (bright) dust slides down a slope and reveals a darker sub-surface.
``Other'' is a catch-all class that contains images that fit none of
the defined classes of interest (e.g., Figure~\ref{fig:hirise-data}(e)). 

We introduce three new classes of interest, which are ``Impact ejecta'', 
``Spider'', and ``Swiss cheese''.
``Impact ejecta'' refers to evidence of a meteorite impact on the
surface.
\comment{
material that is blasted out from the impact of a
meteorite or the eruption of a volcano. We also include cases in which the
impact cleared away overlying dust, exposing the underlying surface. In some
cases, the associated crater may be too small to see. Impact ejecta can also
include lava that spilled out from the impact (blobby ("lobate") instead of
blast-like), more like an eruption (triggered by the impact). Impact ejecta can
be isolated, or they can form in clusters when the impactor breaks up into
multiple fragments.
}
``Spiders'' and ``Swiss cheese'' are phenomena that occur in the south polar
region of Mars.
Spiders have a central pit with radial troughs, and they are believed
to form as a result of seasonal jets expelling carbon dioxide gas
through an overlying ice layer~\cite{aye:polar19}.
%
``Swiss cheese'' is terrain that consists of pits that are formed when
the sun heats the ice making it sublimate (change solid to gas). 

We used a combination of labeling platforms to label the HiRISE
landmark images. 
Early images were labeled by in-house volunteers using the Zooniverse.org
platform.  We conducted a second labeling campaign that targeted three 
minority classes: Impact ejecta, Spiders, and Swiss cheese.  Landmark
images from this campaign were labeled using the \ac{IDAR} browser-based image
labeling tool
\footnote{
\url{https://github.com/stevenlujpl/IDAR}}. 
\comment{
\begin{figure}[ht] 
\centering
\includegraphics[width=8.5cm]{figures/hirise/labeling_tool.png} 
\caption{HiRISE image labeling tool.} 
\label{labelingtool} 
\end{figure}
}
We obtained labels for each image from three volunteers. Images for which the
three labels did not agree (for the second campaign, this amounted to 
approximately 30\% of the images) were manually reviewed to select the
final label.  
To guide labeling when more than one class was present in the image, we
instructed volunteers to prioritize classes
as Impact ejecta, Slope streak, Spider, Dark dune, Bright dune, Swiss
cheese, Crater, or Other.

%


%% file: pc21-msl-dataset.tex
\subsection{MSL Surface Data Set (v2)}

We created a new data set of Mars surface images collected by the
\ac{Mastcam} and \ac{MAHLI} instruments on the \ac{MSL} Curiosity
rover. \ac{Mastcam} is a two-instrument suite with
left- and right-eye cameras. \ac{MAHLI} is a single focusable
camera located on the turret at the end of the rover's robotic arm. In our
previous work~\cite{wagstaff:deepmars18}, we created a data set of \num{6691}
images spanning \num{24} classes that primarily focused on rover hardware
parts. The new data set (v2) includes \num{2900} images spanning \num{19}
classes that primarily focus on objects of scientific
interest~\cite{dataset:msl20}\footnote{\url{https://doi.org/10.5281/zenodo.4033453}}.  

\begin{figure}[t]
\begin{center}
\subfigure[\scriptsize Arm cover]
{\includegraphics[height=0.78in]{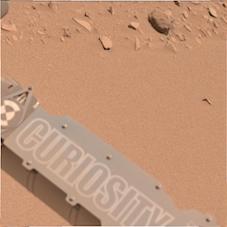}}
\subfigure[\scriptsize Artifact]
{\includegraphics[height=0.78in]{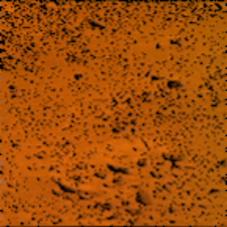}}
\subfigure[\scriptsize Close-up rock]
{\includegraphics[height=0.78in]{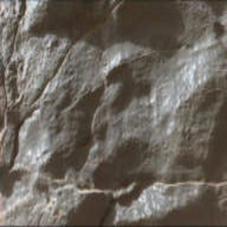}}
\subfigure[\scriptsize Dist. landscape]
{\includegraphics[height=0.78in]{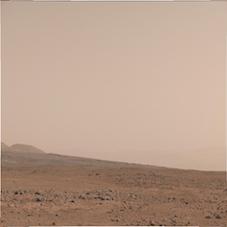}}
\subfigure[\scriptsize Drill hole]
{\includegraphics[height=0.78in]{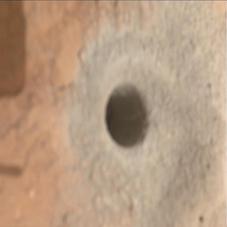}}
\subfigure[\scriptsize DRT]
{\includegraphics[height=0.78in]{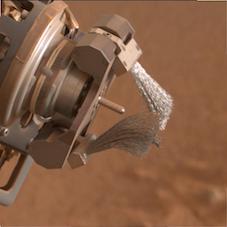}}
\subfigure[\scriptsize DRT spot]
{\includegraphics[height=0.78in]{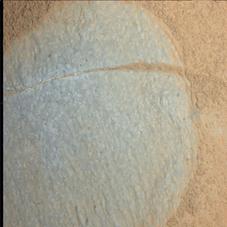}}
\subfigure[\scriptsize Float rock]
{\includegraphics[height=0.78in]{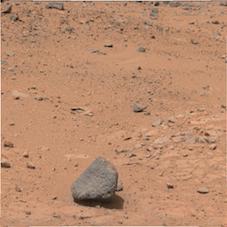}}
\subfigure[\scriptsize Layered rock]
{\includegraphics[height=0.78in]{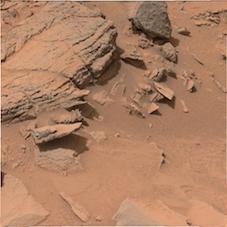}}
\subfigure[\scriptsize L.-toned veins]
{\includegraphics[height=0.78in]{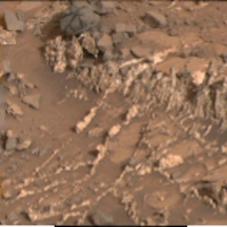}}
\subfigure[\scriptsize M. cal. target]
{\includegraphics[height=0.78in]{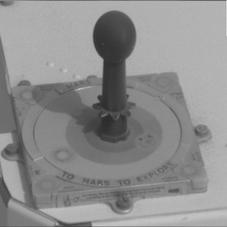}}
\subfigure[\scriptsize Nearby surface]
{\includegraphics[height=0.78in]{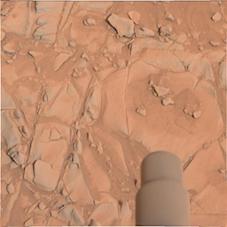}}
\subfigure[\scriptsize Night sky]
{\includegraphics[height=0.78in]{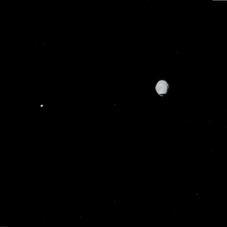}}
\subfigure[\scriptsize O. rover part]
{\includegraphics[height=0.78in]{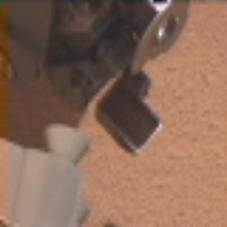}}
\subfigure[\scriptsize Sand]
{\includegraphics[height=0.78in]{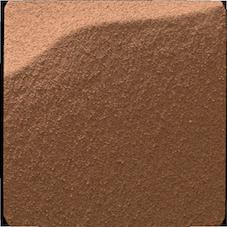}}
\subfigure[\scriptsize Sun]
{\includegraphics[height=0.78in]{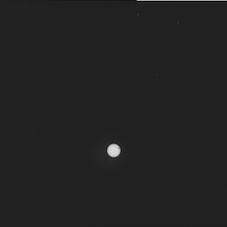}}
\subfigure[\scriptsize Wheel]
{\includegraphics[height=0.78in]{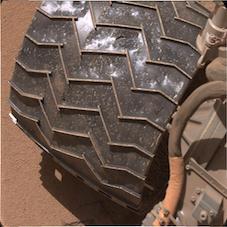}}
\subfigure[\scriptsize Wheel joint]
{\includegraphics[height=0.78in]{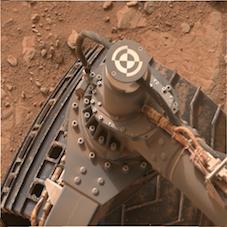}}
\subfigure[\scriptsize Wheel tracks]
{\includegraphics[height=0.78in]{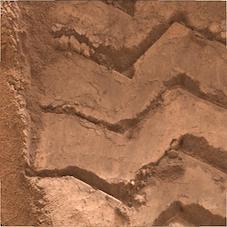}}
\caption{Examples of each class in the \ac{MSL} surface data
  set. Subfigure (d) contains the circular shape of the sun, while
  subfigure (e) is an irregularly shaped moon of Mars.}
\label{fig:msl-ds-v2}
\end{center}
\end{figure}

The \ac{MSL} data set consists of RGB and grayscale images that are
8-bit, decompressed, radiometrically calibrated, color corrected, and
geometrically linearized browse images from the \ac{MSL} mission archive
hosted by the \ac{PDS} \ac{Imaging Node}. 
We resized the smallest side to
\num{227} pixels while preserving aspect ratio
and center-cropped the other side to \num{227} pixels. 
We randomly sampled a total of \num{2900} images from 
sol (\ac{MSL} mission day) \num{1} to \num{2224} composed of 
\num{1172} \ac{Mastcam} left eye camera, \num{1000} \ac{Mastcam} right
eye camera, and \num{728} \ac{MAHLI} images.

Our first task was to define the set of \ac{MSL} classes of interest,
which was not known in advance.
We analyzed a subset of \num{1600} images covering the full sol range
with the browser-based Class Discovery Tool from the \ac{IDAR}
software suite.  This tool allows users to interactively associate
images with dynamically created categories as they are discovered.
We started with an initial list of classes from a domain expert on the
\ac{MSL} science team 
and pre-sorted the images
using the DEMUD algorithm~\cite{wagstaff:demud13} so that the most
``interesting'' or unusual images were displayed first. The
DEMUD algorithm is efficient in terms of class
discovery~\cite{wagstaff:realml20}. The class discovery process yielded
\num{19} classes of interest.

Examples of each class are shown in
Figure~\ref{fig:msl-ds-v2} in alphabetical order. They include three classes 
that describe rover-created features
(``Drill hole'',``DRT (Dust Removal Tool) spot'', and ``Wheel tracks''),
``Sun'' and ``Night sky'', seven Mars surface feature classes
(e.g., ``Light-toned veins'', ``Layered rock'', ``Float rock''),
five rover part classes (e.g., ``DRT'', ``Mastcam calibration target'',
``Wheel'' and a generic ``Other rover part'' class), and 
``Artifact'' used for images that are low in quality or contain
missing data.  
The pixel resolutions and lighting conditions in these
images vary a lot as they were imaged at different distances
and different times.

We used the \ac{IDAR} labeling tool to
label \ac{MSL} surface data set images. We divided the \num{2900}
images into \num{29} batches, and each batch of \num{100} images
were distributed to three volunteers for labeling. 
We provided detailed labeling instructions with
definitions of each class and prioritization guidance when
multiple classes appeared in a single image. 
As with the HiRISE data set, we resolved disagreement using expert
review. 
The \ac{MSL} surface data set class distribution 
is shown in Table~\ref{tab:msl-ds-dist}.

\begin{table}
\centering
\begin{tabular}{l|ccc} 
\bf Class Name & \bf Count & \bf Percent \\ \hline
Arm cover & \num{10} & 0.34\% \\ 
Artifact & \num{408} & 14.07\% \\ 
Close-up rock & \num{373} & 12.86\% \\ 
Distant landscape & \num{197} & 6.79\% \\ 
Drill hole & \num{65} & 2.24\% \\ 
DRT & \num{14} & 0.48\% \\ 
DRT spot & \num{47} & 1.62\% \\ 
Float rock & \num{80} & 2.76\% \\ 
Layered rock & \num{105} & 3.62\% \\ 
Light-toned veins & \num{42} & 1.45\% \\ 
Mastcam calibration target & \num{100} & 3.45\% \\ 
Nearby surface & \num{1008} & 34.76\% \\ 
Night sky & \num{23} & 0.79\% \\ 
Other rover part & \num{86} & 2.97\% \\ 
Sand & \num{123} & 4.24\% \\ 
Sun & \num{115} & 3.97\% \\ 
Wheel & \num{56} & 1.93\% \\ 
Wheel joint & \num{33} & 1.14\% \\ 
Wheel tracks & \num{15} & 0.52\% \\ \hline
Total & \num{2900} & 100\% \\ \hline
\end{tabular}
\caption{MSL (Mars surface) data set class distribution.}
\label{tab:msl-ds-dist}
\end{table}


%% file: pc21-classifier.tex
\section{CNN Classification for Mars Images}

\begin{table*}
\centering
\begin{tabular}{l|cc|cc|cc|cc}
{} & 
\multicolumn{2}{c|}{\bf Train sol 1 - 948} & 
\multicolumn{2}{c|}{\bf Val. sol 949 - 1920} & 
\multicolumn{2}{c|}{\bf Test sol 1921 - 2224} &
\multicolumn{2}{c}{\bf Full Archive sol 1 - 2224} \\ 
\bf Instrument & \bf Count & \bf Percent & \bf Count & \bf Percent & \bf Count 
& \bf Percent & \bf Count & \bf Percent \\ \hline
Mastcam Left & \num{842} & 42.1\% & \num{108} & 36.0\%
             & \num{222} & 37.0\% & \num{50480} & 40.0\% \\ 
Mastcam Right & \num{678} & 33.9\% & \num{107} & 35.7\%
              & \num{215} & 35.8\% & \num{43041} & 34.1\% \\ 
MAHLI & \num{480} & 24.0\% & \num{85} & 28.3\%
      & \num{163} & 27.2\% & \num{32612} & 25.9\% \\ \hline
Total & \num{2000} & 100\% & \num{300} & 100\%
      & \num{600} & 100\% & \num{126133} & 100\% \\ 
\hline
\end{tabular}
\caption{MSL surface data set training, validation, and test data sets
by instrument and sol range.}
\label{tab:msl-inst-sol-distribution}
\end{table*}

We trained and deployed two \ac{CNN} classifiers, denoted 
as HiRISENet and MSLNet, for \ac{MRO} \ac{HiRISE} images and \ac{MSL} 
\ac{Mastcam} and \ac{MAHLI} images.
We  
employed transfer learning to adapt the weights of networks pre-trained 
on Earth images for use with Mars orbital and surface images. 

\input{pc21-hirisenet}
\input{pc21-mslnet}


%% file: pc21-hirisenet.tex
\subsection{HiRISENet: CNN Classifier for Mars Orbital Images} 

We adapted the AlexNet image
classifier~\cite{krizhevsky:alexnet12} for use with HiRISE classes.
AlexNet was trained on 1.2 million Earth images from 1000 classes in the
ImageNet data set. 
We started with Caffe's BVLC reference model~\cite{jia:caffe14}, which
is a copy of AlexNet 
that was trained for \num{310000} iterations, provided by Jeff
Donahue\footnote{https://github.com/BVLC/caffe/tree/master/models/bvlc\textunderscore
reference \textunderscore caffenet}. 
We removed the final fully connected layer, added a new layer with
eight output classes, and re-trained the network with
Caffe~\cite{jia:caffe14}. We followed Caffe's recommendations for
fine-tuning, including using a small base learning rate and small step
size and a larger learning rate multiplier for the final layer
only. We used a learning rate of 0.0001, weight decay of
0.0005, and relatively small step size of \num{20000}. The 
initial layers were almost fixed; they used learning rate multipliers
of 1 (weight) and 2 (bias).  The final layer was allowed greater
adaptation with multipliers of 10 (weight) and 20 (bias).  We
trained the model for \num{78900} iterations.
Caffe computes the per-band mean pixel values from the training set
and uses these values to normalize all images during training and
prediction. 

We split the HiRISE dataset into train, validation, and test sets 
using each landmark's HiRISE source image identifier to ensure
no overlap in 
source images between the sets. We used approximately 65\% of
the data for training, 19\% for validation, and 17\% for
testing. Images obtained from our second labeling campaign (to target
minority classes) appear in the training and validation sets only so
that we could assess improvements against an unchanged test set.

We applied data augmentation to the training and validation sets. The
augmentation includes three rotations: 90, 180, and 270 degrees,
horizontal and vertical flips, and a random brightness adjustment. In
addition, we up-sampled data obtained in the second labeling campaign
by a factor of two.


%% file: pc21-mslnet.tex
\subsection{MSLNet: CNN Classifier for Mars Surface Images}

MSLNet is a hybrid of two \ac{CNN} classifiers.
The \ac{v1} classifier 
focused on rover hardware classes~\citep{wagstaff:deepmars18, 
lu:pdw19}. The primary objectives of the \ac{Mastcam} and \ac{MAHLI} 
instruments are to enable science analysis of rover investigation
sites, which motivated the creation of \ac{v2} classifier to expand 
the set of classes to include science targets (e.g. ``Float rock'', ``Layered 
rock'') and activities (e.g. ``DRT spot'', ``Drill hole''). The \ac{v1} 
classifier initially focused on engineering considerations and rover hardware
classes due to requests by the \ac{MSL} rover planning team as well as 
pre-existing availability of labels for those items. The ``Wheel'' class was
of particular interest due to growing awareness in 2017 that the rover's 
wheels were experiencing a higher than expected level of degradation due to
driving on the rough surface. The success of the \ac{v1} classifier led to new
requests to also accommodate science-related classes in support of \ac{MSL}
mission to explore and understand Mars. Observations that contain classes such 
as ``Layered rock'' and ``Light-toned veins'' are very high science priorities
to help determine the history and evolution of water activity, which can also 
have implications for habitability. The deployed MSLNet classifier covers both
areas of interest (engineering and science) to meet the needs of diverse users
with different priorities.

MSLNet 
first classifies images with the \ac{v2} classifier, then reclassifies
any images classified as ``Other rover part'' with the \ac{v1} classifier 
to get a fine-grained classification of rover parts. The creation and 
evaluation of the \ac{v1} classifier were reported in 
previous work~\cite{wagstaff:deepmars18}. 
The \ac{v2} classifier was trained and evaluated using the MSL surface
data set described above. 
We divided this data set into training, validation, and test data 
sets according to their sol of acquisition to 
enable the evaluation of generalization performance on newly acquired
images. 
The sol splitting boundaries, as shown in
Table~\ref{tab:msl-inst-sol-distribution}, were chosen to enable 
per-camera distributions that roughly match the full archive.

To improve the generalization performance of the classifier, we 
augmented the images in the training data set (but not 
the validation and test data sets). The \ac{MAHLI} images (which come
from a rotatable platform) were 
augmented using rotation ($90^{\circ}$, $180^{\circ}$, and
$270^{\circ}$) and flipping (horizontal and vertical); 
the \ac{Mastcam} images (which come from a fixed platform) were
augmented using only horizontal and vertical flipping methods. 
\comment{
The  
\ac{MAHLI} and \ac{Mastcam} images were augmented differently because the 
locations on which the instruments are mounted. The \ac{MAHLI} instrument is 
mounted on the rotatable turret of the rover's robotic arm, whereas the 
\ac{Mastcam} instrument is mounted on the rover's Remote Sensing Mast that can not 
be rotated. 
}

As with HiRISENet, for the MSLNet \ac{v2} classifier we fine-tuned AlexNet
for \num{2050} iterations with a fixed base learning rate of $0.0001$. We set 
the learning rate multipliers of the first four convolution layers, the fifth 
convolution layers, and the final fully connected layers to 0, 1, 20 
respectively. We set the dropout rate to $0.5$ for the first and second fully 
connected layers.
The final hybrid MSLNet classifier combines \ac{v1} and {v2} and 
classifies images into \num{35} classes of both science and engineering 
relevance.


%% file: pc21-calibration.tex
\section{Classifier Calibration} 

The deployed classifiers use a confidence threshold
to determine which results are shown to users, so it is vital 
that the models are well calibrated. Modern neural networks 
have achieved higher accuracies but in many
cases have suffered an increase in calibration error, which means that
the predicted class probabilities deviate from the true empirical
probabilities.  In many cases, the networks are consistently
over-confident in their predictions.  This effect appears to
be tied to an increase in model capacity and lack of
regularization~\cite{guo:calib17}. 
For a quantitative measure of model calibration, we
calculate the Expected Calibration Error (ECE),
which is the expected difference 
between posterior probability (confidence) and accuracy.  We
partition $n$ predictions into $M$ equally spaced bins and
computed a population-weighted average of the difference between
accuracy and confidence within each bin: \\
\centerline{
$\text{ECE} = \sum_{m=1}^M\frac{|B_m|}{n} |\text{acc}(B_m) -
\text{conf}(B_m)|.$  
}

\comment{
The Maximum Calibration Error computes the maximum deviation between
accuracy and confidence across all $M$ bins:

\begin{equation} 
\label{eqn:mce} 
\text{MCE} = \max_{m\in\{1,...,M\}} |\text{acc}(B_m) - \text{conf}(B_m)| 
\end{equation}
}

We evaluated four post-hoc calibration methods that
extend Platt Scaling~\cite{platt1999probabilistic} to
multiclass problems. 
%
Temperature scaling~\cite{guo:calib17} uses
a single parameter $T$ to rescale model output.  Given the model
output for item $\mathbf{x}$, which is a logit vector $\mathbf{z} \in
\mathcal{R}^K$, 
the calibrated probability of class $k$ is:
$  p(y=k|x)=\frac{e^{z_k/T}}{\sum_{j=1}^K{e^{z_j/T}}}$.
%
The parameter $T$ is optimized with respect to the log likelihood on
the validation set. Since the parameter $T$ does not change the
maximum of the softmax function, the accuracy of the model is
unchanged.
Bias-corrected temperature scaling (BCTS)~\cite{alexandari:bcts20}
adds a bias term $b_k$ for each class:
 $ p(y=k|x)=\frac{e^{z_k/T + b_k}}{\sum_{j=1}^K{e^{z_j/T + b_j}}}$.
%
Vector and matrix scaling~\cite{guo:calib17} add additional flexibility
with per-class scaling using a $K \times K$ linear transformation
matrix $\mathbf{W}$ by computing $\mathbf{W} \mathbf{z} + \mathbf{b}$
and then normalizing across classes to get $p(y=k|x)$. Vector scaling
constrains $\mathbf{W}$ to be a diagonal matrix whose entries function
as class-specific temperature values.


%% file: pc21-evaluation.tex
\section{CNN Classification Evaluation}

To evaluate HiRISENet and MSLNet,
we used the overall (post-threshold) accuracy score and abstention rate as
the primary performance metrics and ECE to 
measure the calibration level of the classifiers. 
We also analyzed the precision and recall scores to understand 
per-class performance. 

\input{pc21-hirisenet-eval}
\input{pc21-mslnet-eval}


%% file: pc21-hirisenet-eval.tex
\subsection{HiRISENet Evaluation}

HiRISENet classification accuracy results are shown in
Table~\ref{tab:hirise}.  Random class prediction on this data set
achieves 11.1\% accuracy (given eight classes).
Compared to a simple baseline that predicts the most common class
from the training set (``Other''), HiRISENet exhibits a strong
improvement from 81.1\% to to 92.8\% on the test set.

\comment{
HiRISENet outperforms both baselines. The test set accuracy is higher
than our validation set accuracy, but the test set has a larger
percentage of the ``Other'' class.
}

\begin{table}
\begin{center}
\begin{tabular}{l|ccc}
            & Train & Val & Test \\ \
	    & \small{($n=6997$)} & \small{($n=2025$)} & \small{($n=1793$)} \\ 
Classifier  & \tiny{($n_{aug}=51058$)} & \tiny{($n_{aug}=14959$)} &  \\ \hline 
Most common & 78.4\% & 75.0\% & 81.1\% \\  \hline
HiRISENet & {\bf 99.6\%} & {\bf 88.6\%} & {\bf 92.8\%} \\ \hline

\end{tabular}
\end{center}
\caption{Classification accuracy on HiRISE (Mars orbital) images.
The best performance on each data set is in bold.}
\label{tab:hirise}
\end{table}

\begin{table}
\begin{center}
\comment{
\begin{tabular}{l|ccc|cc}
        &      &     &     & \multicolumn{2}{c}{0.9 confidence} \\
        &  ECE & MCE & Acc & Acc & Abst Rate  \\ \hline
        Un-calibrated  & 0.073 & 0.032 & 88.6  & 94.2 & \bf{13\%} \\ \hline
Temperature & 0.013 & 0.002 & 88.6 & 97.3 & 31\% \\
Temp w/Bias & 0.014 & 0.002 & 89.2 & 97.3 & 29\% \\
        Vector Scale & \bf{0.010} & \bf{0.002} & 89.3 & 97.2 & 27\% \\
        Matrix Scale & 0.013 & 0.002 & \bf{90.3} & \bf{97.7} & 24\% \\ \hline
}
\begin{tabular}{l|cc|cc}
        &      &     & \multicolumn{2}{c}{0.9 confidence} \\
        &  ECE & Acc & Acc & Abst Rate  \\ \hline
        Uncalibrated  & 0.073 & 88.6  & 94.2 & \bf{13\%} \\ \hline
Temperature scaling & 0.013 & 88.6 & 97.3 & 31\% \\
BCTS & 0.014 & 89.2 & 97.3 & 29\% \\
        Vector scaling & \bf{0.010} & 89.3 & 97.2 & 27\% \\
        Matrix scaling & 0.013 & \bf{90.3} & \bf{97.7} & 24\% \\ \hline
        \end{tabular}
\end{center}
\caption{HiRISENet calibration results on validation set; best
  performance in bold.}  
\label{tab:ece-mce}
\end{table}

\comment{
The accuracy results shown so far are without classifier
calibration. In attempts to make our classifier's predicted
probabilities closer to real probabilities, we explore the four
calibration methods discussed in the previous section.}
For this application, reliable posterior probabilities are essential,
since predictions are thresholded so that only those of at
least 0.9 probability are shown to users.  We compared four
calibration methods in terms of their impact on accuracy, ECE, and
abstention (Table~\ref{tab:ece-mce}).
We found that matrix scaling achieved the highest validation accuracy
($90.3$\%) as well as the lowest abstention rate ($24$\%).
Vector scaling achieved the lowest ECE but with higher abstention and
lower accuracy.  Therefore, we adopted matrix scaling for deployment.
On the test set, the calibrated HiRISENet model achieved $96.7$\%
accuracy with an abstention rate of $20$\%.

\comment{
\begin{table}
\begin{center}
\begin{tabular}{l|ccc}
HiRISENet (.90 conf)  & Train & Val & Test \\ \hline
Abstention rate   & 13\% & 24\% & 20\% \\
Accuracy & 99.5\% & 97.7\% & 96.7\% \\  \hline
\end{tabular}
\end{center}
\caption{Classification accuracy on HiRISE (Mars orbital) images, at
  equal to or greater than 90\% confidence. Model Calibrated with
  matrix scaling.} 
\label{tab:hirise_acc_abs}
\end{table}

In Table~\ref{tab:hirise_acc_abs}, we show a comparison of the
training, validation, and test set accuracies and abstention rates for
our model calibrated with matrix scaling. We see that when the
classifier is confident, it has very high accuracy. The abstention
rate for the test set overall is 20\%. On a per class basis, most
classes had abstention rates between 15 and 40\%, except dark dunes
had a rate of almost 80\%.
}

\begin{figure}[!tbp] 
\centering
\includegraphics[width=8.5cm]{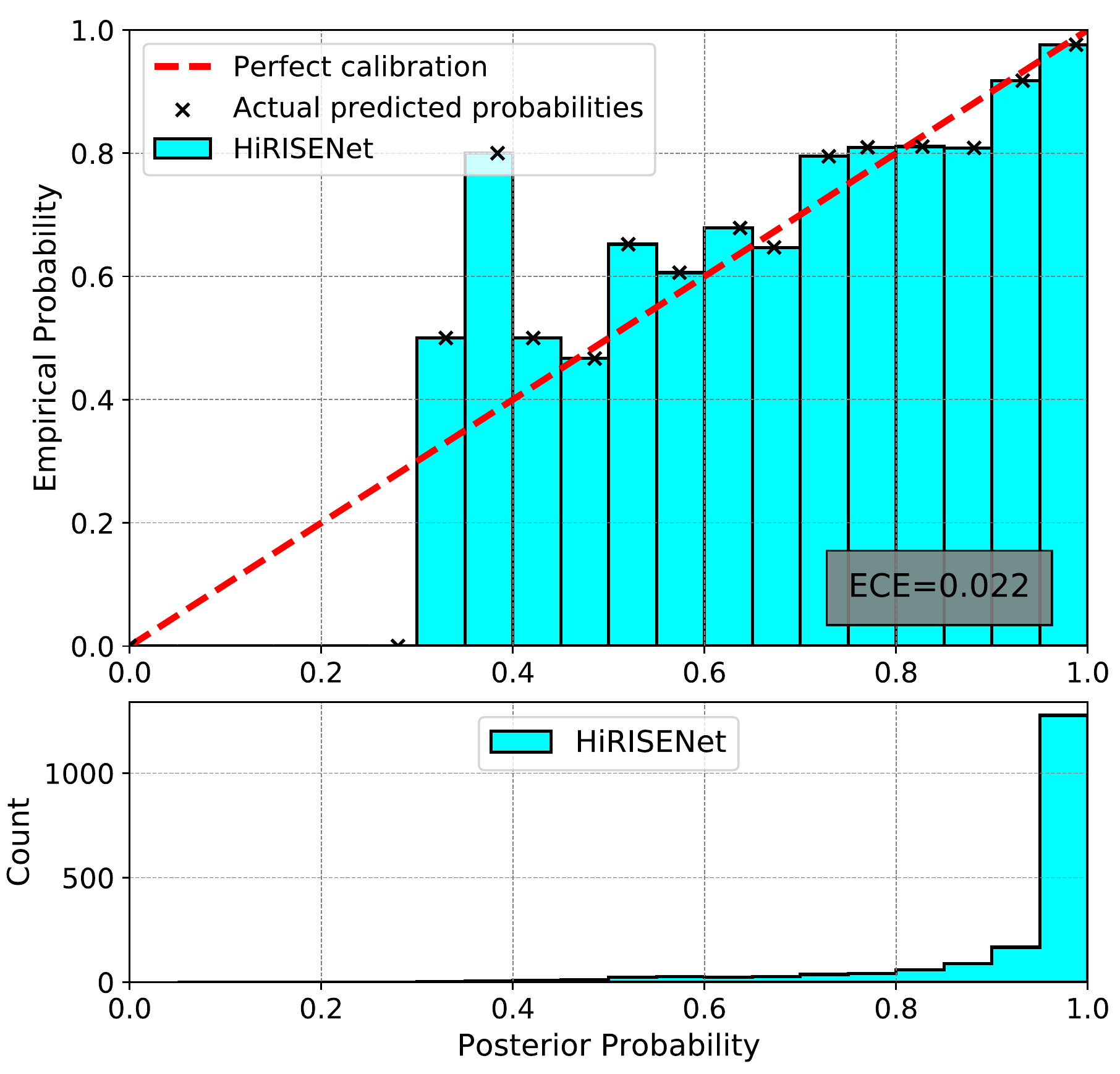}
\caption{Calibrated HiRISENet reliability (test set).}
\label{ece_matrix}
\end{figure}

Reliability
diagrams~\cite{degroot1983comparison,niculescu2005predicting} provide
a visual representation of model calibration.  The empirical per-bin
accuracy is plotted as a function of model posterior
probability.  For a perfectly calibrated model, these values are equal,
following the diagonal line.
Figure~\ref{ece_matrix} shows the reliability diagram for HiRISENet.
The bottom panel shows the data set distribution in terms of predicted
probability.  We find that HiRISENet is well calibrated with an ECE of
just $0.022$ and the majority of predictions in the most-confident bin.

\comment{
The abstention rate per class is shown in Figure~\ref{test-abst}. The green
circle shows the point at which confidence is at 90 \%. We see that ``Spiders''
show decreasing accuracy with increased abstention rate, and there are regions
where ``impact ejecta'' and ``crater'' have decreasing accuracy with increasing
abstention rate. Overall, however, we see the tend of increasing accuracy with
increasing abstention. 
\begin{figure}[ht]
\centering
\includegraphics[width=8.5cm]{figures/hirise/abst_class_matrix.pdf}
\caption{HiRISE Classifier with Matrix Scaling: Abstention Rate per Class for
Test Set.} 
\label{test-abst} 
\end{figure}
}


\begin{figure}[t]
\centering
\includegraphics[width=8.5cm]{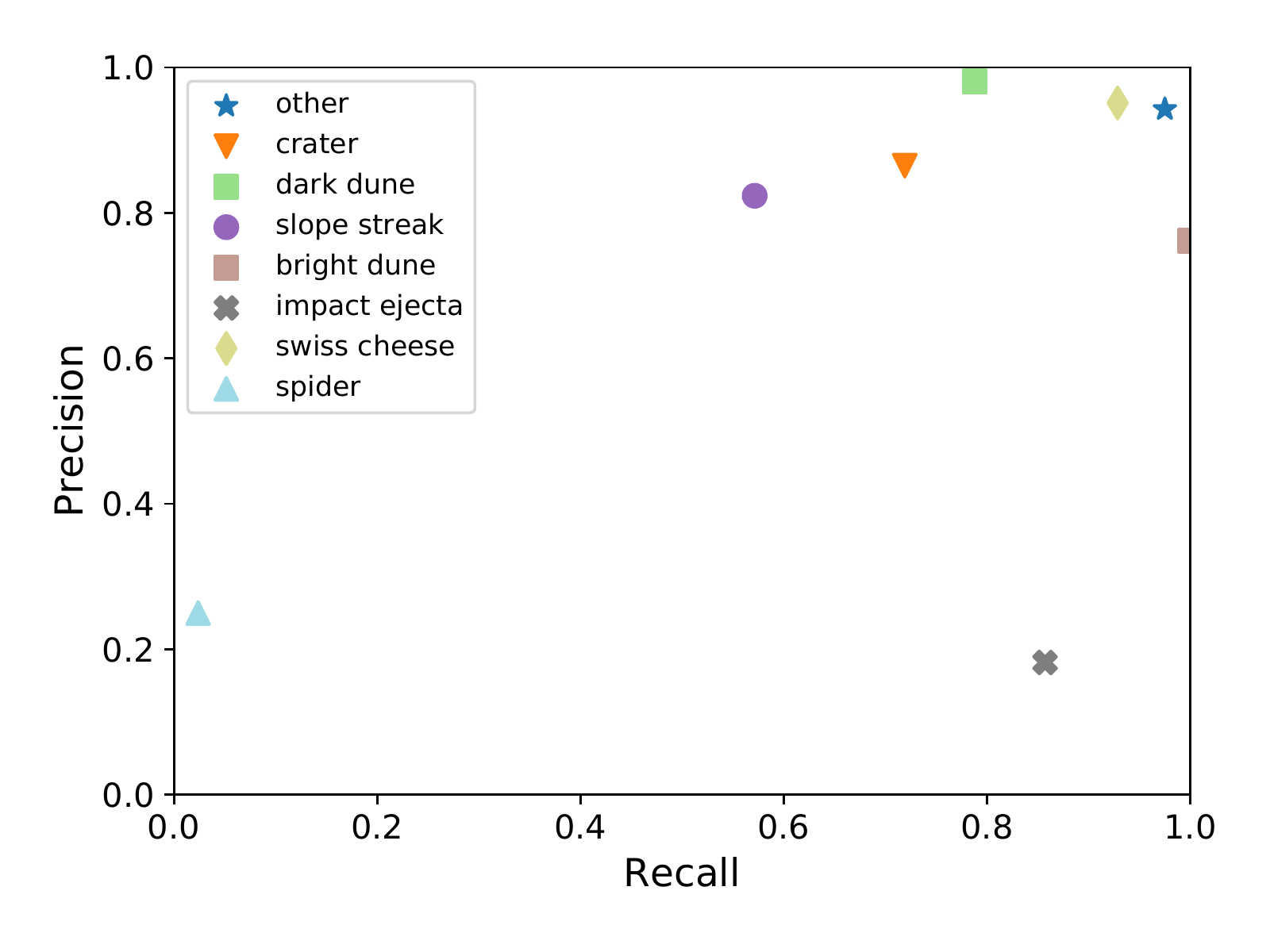}
\caption{Calibrated HiRISENet per-class precision and recall (test set).}
\label{pr-test-hirise}
\end{figure}

\begin{figure}[t]
\centering
\includegraphics[width=8.5cm]{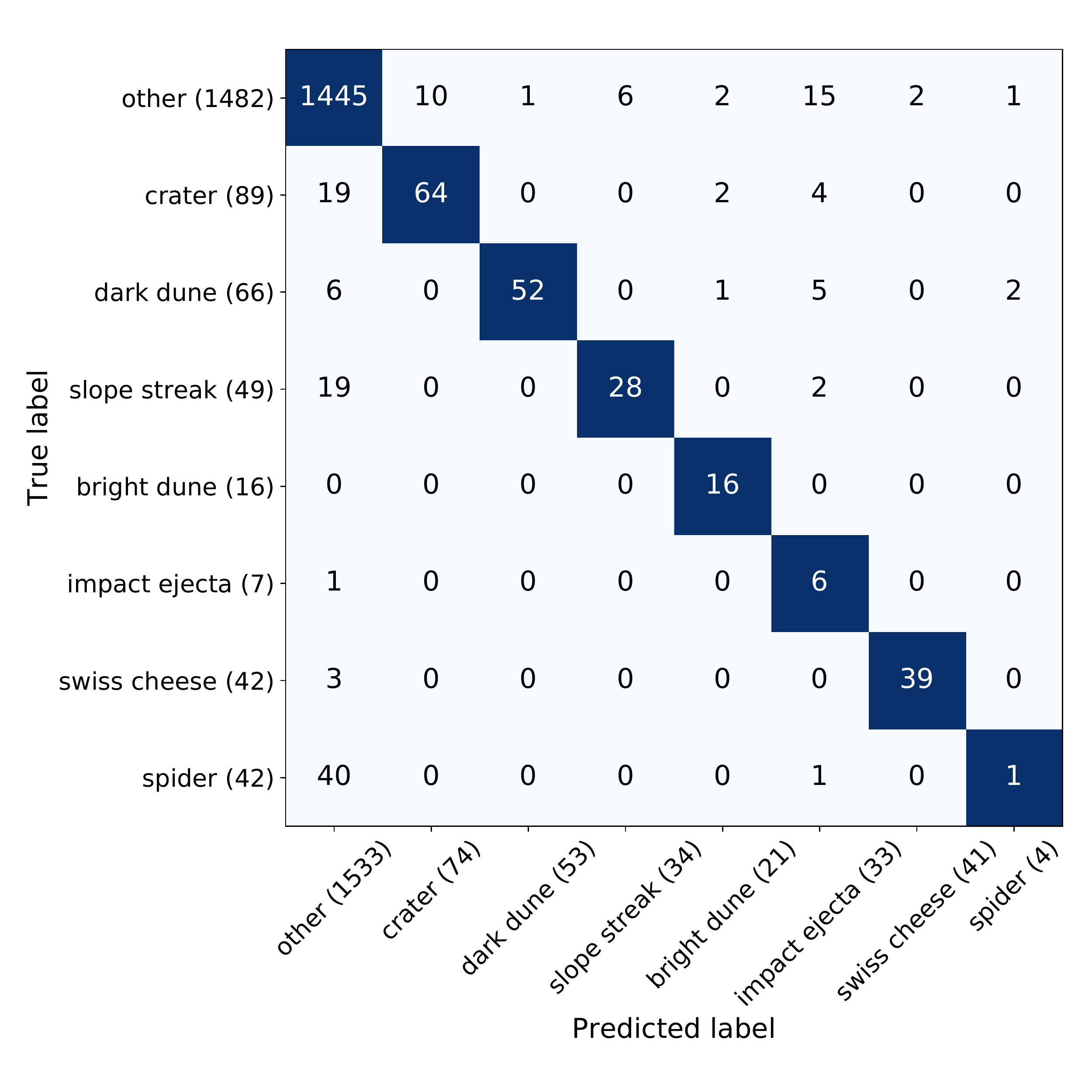}
\caption{Calibrated HiRISENet confusion matrix (test set).}
\label{cm-test-hirise}
\end{figure}

\begin{figure}[t]
\centering
\subfigure[Validation set]{
\includegraphics[width=3.9cm]{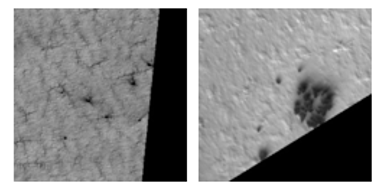}
\label{vs-spiders}
}
\subfigure[Test set]{
\includegraphics[width=3.9cm]{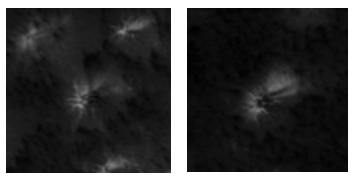}
\label{ts-spiders}
}
\caption{Domain shift in HiRISE ``Spider'' landmarks.}
\label{fig:shift}
\end{figure}

Figure~\ref{pr-test-hirise} shows per-class precision and recall on
the test set after matrix scaling calibration.
Most classes achieve precision above $0.70$ with recall above $0.50$
(even after thresholding).
The ``Spider'' class has the lowest recall ($0.02$, out of only $42$
items), while
the ``Impact ejecta'' class has the lowest precision ($0.18$).
Figure~\ref{cm-test-hirise} shows the confusion matrix on the test set 
after matrix scaling calibration.  Diagonal (correct) entries have a
dark background.  A comparison to the confusion matrix
before calibration (not shown) indicates that two images that were
previously incorrectly classified into the ``Spider'' class are now
correctly classified as the ``Impact ejecta'' class; however,
the confusion between the ``Crater'' and ``Impact ejecta'' classes has
increased. 
%
In addition, the ``Spider'' class suffered from significant domain shift, which
is evident in Figure~\ref{fig:shift}. The ``Spider'' images in the validation 
set as shown in Figure~\ref{fig:shift}(a) are extremely visually different 
compared to the ``Spider'' images in the test set as shown in 
Figure~\ref{fig:shift}(b). We found that even human labelers had trouble 
recognizing them as the same phenomena. Future updates to this data set will
target the ``Spider'' class.


%% file: pc21-mslnet-eval.tex
\subsection{MSLNet Evaluation}

\begin{table*}
\centering
\begin{tabular}{l|c @{\hspace{0.2cm}} c @{\hspace{0.2cm}} c|c @{\hspace{0.2cm}} c @{\hspace{0.2cm}} c|c @{\hspace{0.2cm}} c @{\hspace{0.2cm}} c}
{} & 
\multicolumn{3}{c|}{\bf Train (n=5920)} & 
\multicolumn{3}{c|}{\bf Validation (n=300)} & 
\multicolumn{3}{c}{\bf Test (n=600)} \\ 
{} & Acc & Acc (0.9) & Abst Rate & Acc & Acc (0.9) & Abst Rate & Acc & Acc (0.9) & Abst Rate \\ \hline
Most Common & 26.3\% & - & - & 24.7\% & - & - & 31.2\% & - & - \\ \hline 
MSLNet & \bf 99.6\% & \bf 100\%  & \bf 6.5\% & \bf 78.3\%  & 89.8\% & \bf 38.0\% & \bf 74.5\%  & 87.2\% &
	\bf 36.2\% \\
	\hline
MSLNet-calibrated & \bf 99.6\%  & \bf 100\% & 18.8\% & \bf 78.3\% & \bf 96.5\% & 52.7\% &
	\bf 74.5\% & \bf 90.3\% & 51.8\% \\ \hline
 \comment{
MSLNet-squeezenet & 98.53\% & 25.39\% & 92.44\% & 42.67\% & 87.76\% & 44.17\% \\
\hline
MSLNet-squeezenet-calibrated & 99.79\% & 43.11\% & 94.02\% & 61.00\% &
	88.98\% & 59.17\% \\ \hline
}
\end{tabular}
\caption{Performance results for MSLNet classifiers (best for each data set is 
	in bold). Note that Acc (0.9) in the title means accuracy score computed 
	with a 0.9 confidence threshold.}
\label{tab:mslnet-perf}
\end{table*}

The performance of the MSLNet \ac{v2} classifier is shown in 
Table~\ref{tab:mslnet-perf} in comparison to the most-common-class
(``Nearby surface'') baseline.
The MSLNet classifier significantly outperforms the baseline
method, achieving $74.5$\% accuracy, or $87.2$\% with $36$\% abstention using a confidence threshold of $0.9$, on the test
set. 
\comment{
To summarize  
the performance, the training set accuracy score and abstention rate are $99.58
$\% and $6.49$\%; the validation set accuracy score and abstention rate are $
89.78$\% and $38.00$\%; the test set accuracy score and abstention rate are $
87.21$\% and $36.17$\%. 
}
%
Recall that images 
in the training, validation, and test sets were divided according to their sol 
(date) of acquisition. The performance of the classifier gradually
decreases over time as the rover traversed to new locations,
possibly due to label shift, 
in which the prior class probabilities change over space or
time~\cite{lipton:labelshift18}.  We plan to investigate 
label shift adaptation to enable the classifier to accommodate such
change. 

MSLNet achieves lower accuracy and higher abstention than HiRISENet on
its corresponding test set.  
Given the larger number of classes and smaller number of 
labeled images, we believe that this classifier is likely even more
data-limited 
and would benefit from additional data collection.

\begin{figure}[t]
\centering
\includegraphics[width=8.5cm]{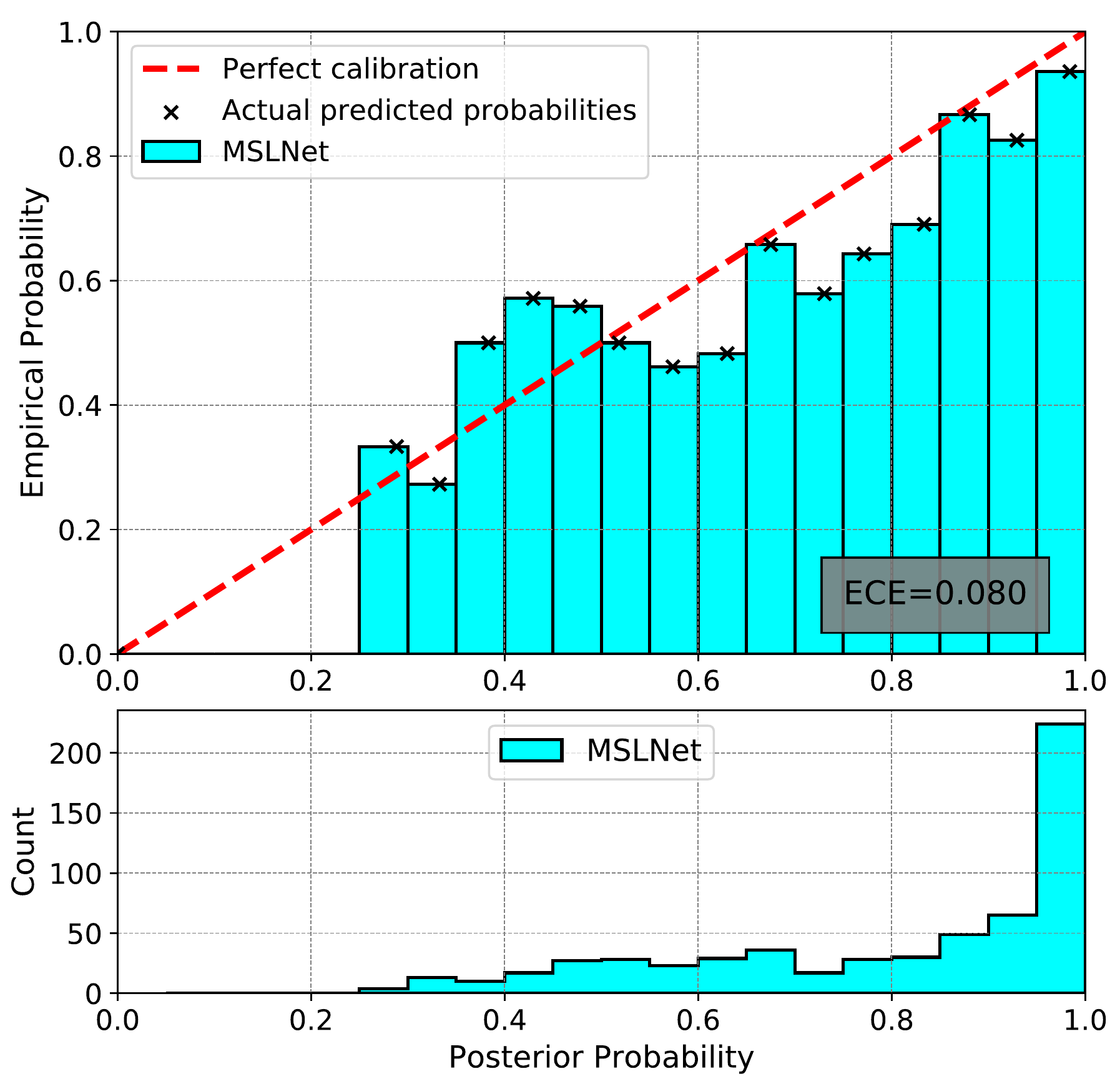}
\caption{Calibrated MSLNet reliability diagram (test set).}
\label{fig:mslnet-reliability-diagram}
\end{figure}

Reliable posterior probabilities are likewise essential for MSLNet.
We calibrated the MSLNet classifier using 
temperature scaling, the most computationally efficient method 
among the four calibration methods discussed in this paper (e.g.,
matrix scaling scales quadratically with the number of classes, which
is problematic for MSLNet).
After calibration, test set accuracy using the confidence threshold
improved to $90.3$\%, at the cost of increased abstention.  For this 
application, we are willing to sacrifice coverage to ensure that the
classifications provided to users are highly reliable.
MSLNet's ECE improved from $0.142$ to $0.08$ with temperature
scaling. The reliability diagram of MSLNet after calibration is  
shown in Figure~\ref{fig:mslnet-reliability-diagram}.
 
\begin{figure}
\centering
\includegraphics[width=8.5cm]{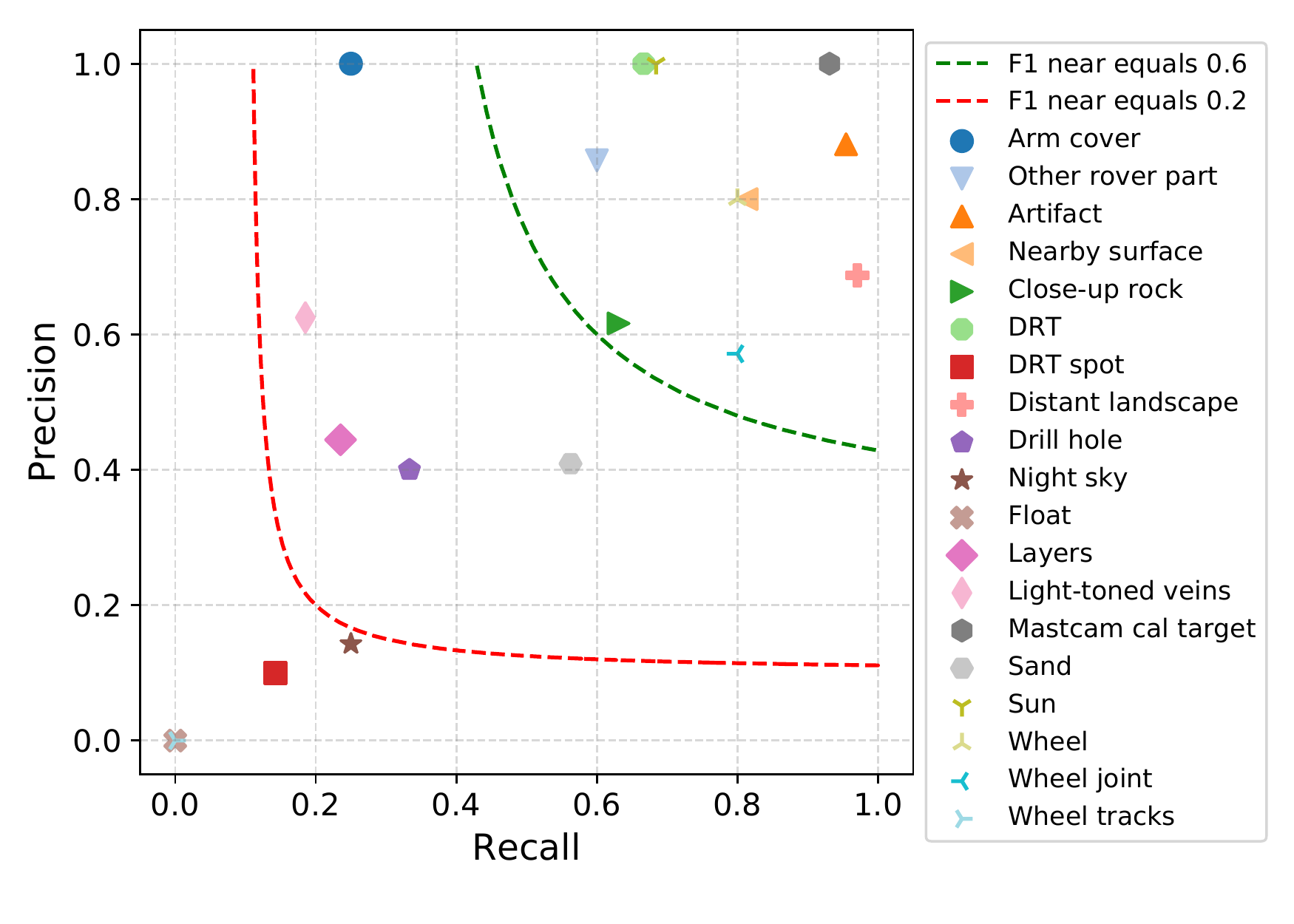}
\caption{Calibrated MSLNet per-class precision and recall (test set).}
\label{fig:mslnet-pr}
\end{figure}

Per-class precision and recall scores are shown 
in Figure~\ref{fig:mslnet-pr}. The green and red F1-score curves separate the 
classes into three groups. The first group, those
above the green curve, includes ten classes (e.g. ``Nearby surface'',
``Mastcam calibration target'') whose F1 scores are greater than
$0.6$; the second (intermediate) group
includes five classes (e.g. ``Layered rock'', ``Drill hole'') whose F1 scores 
are between $0.2$ and $0.6$; and the third group, those
below the red curve, includes four classes (e.g. ``Float rock'', 
``Wheel tracks'') whose F1 scores are less than $0.2$. We note that the 
classes in the third group were evaluated on very few images, so 
their performances are not statistically robust. These classes require
further improvement, and we plan to investigate up-sampling or
additional data acquisition to increase the number of images of these classes.

%% file: pc21-deployment.tex
\section{PDS Image Atlas Deployment}

\begin{figure}
\centering
\includegraphics[width=2.5in]{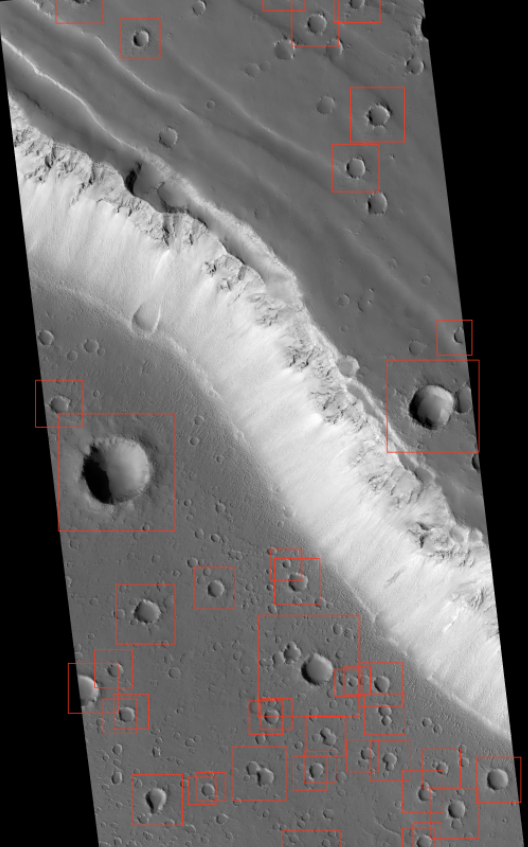}
\caption{View from the Atlas of craters found in HiRISE image
  PSP\_002912\_2075\_RED.} 
\label{fig:atlas}
\end{figure}

\begin{figure*}[h]
\centering
\subfigure[HiRISENet queries]
{\includegraphics[height=2.5in]{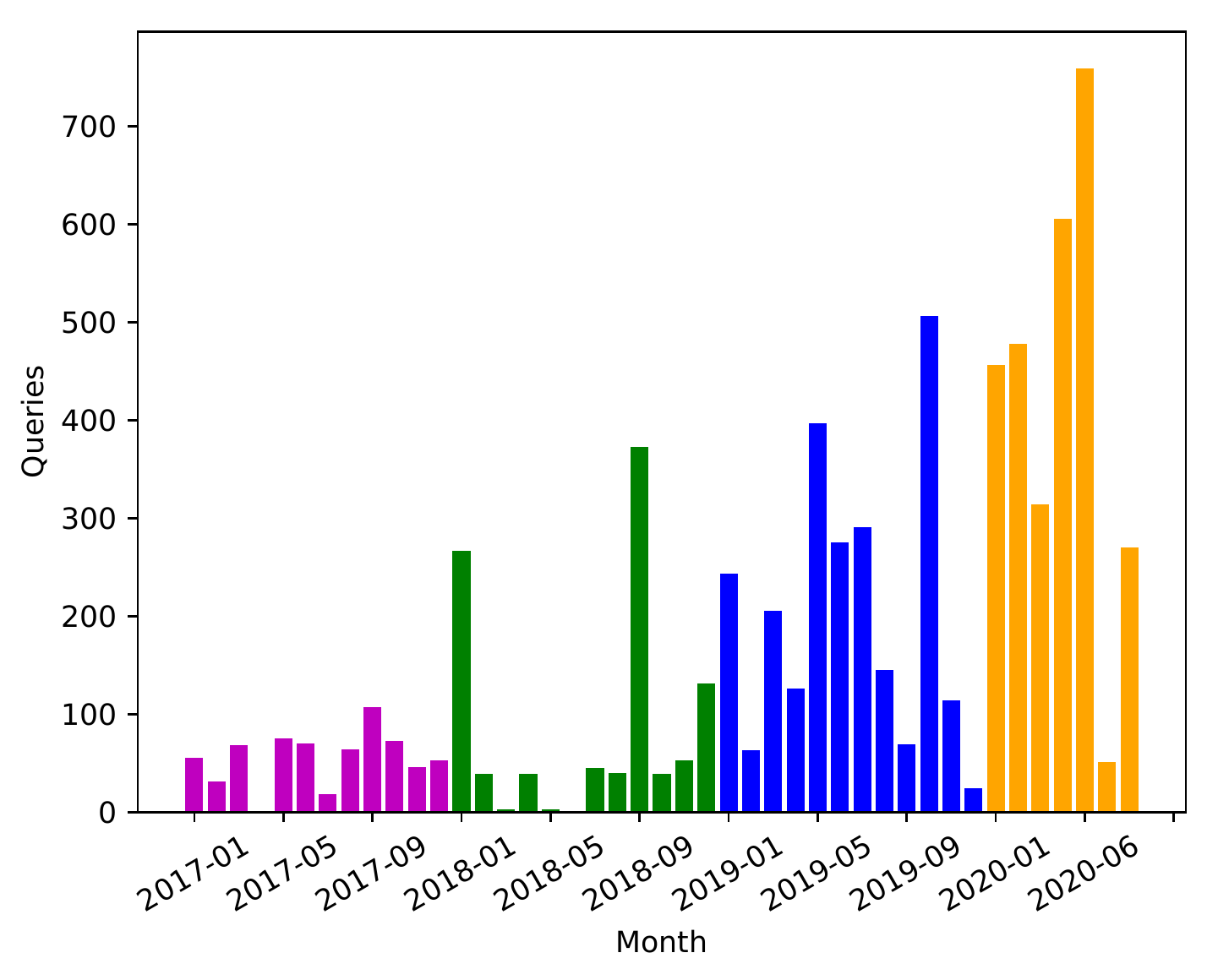}
\label{fig:mrousage}
}
\subfigure[MSLNet queries]
{\includegraphics[height=2.5in]{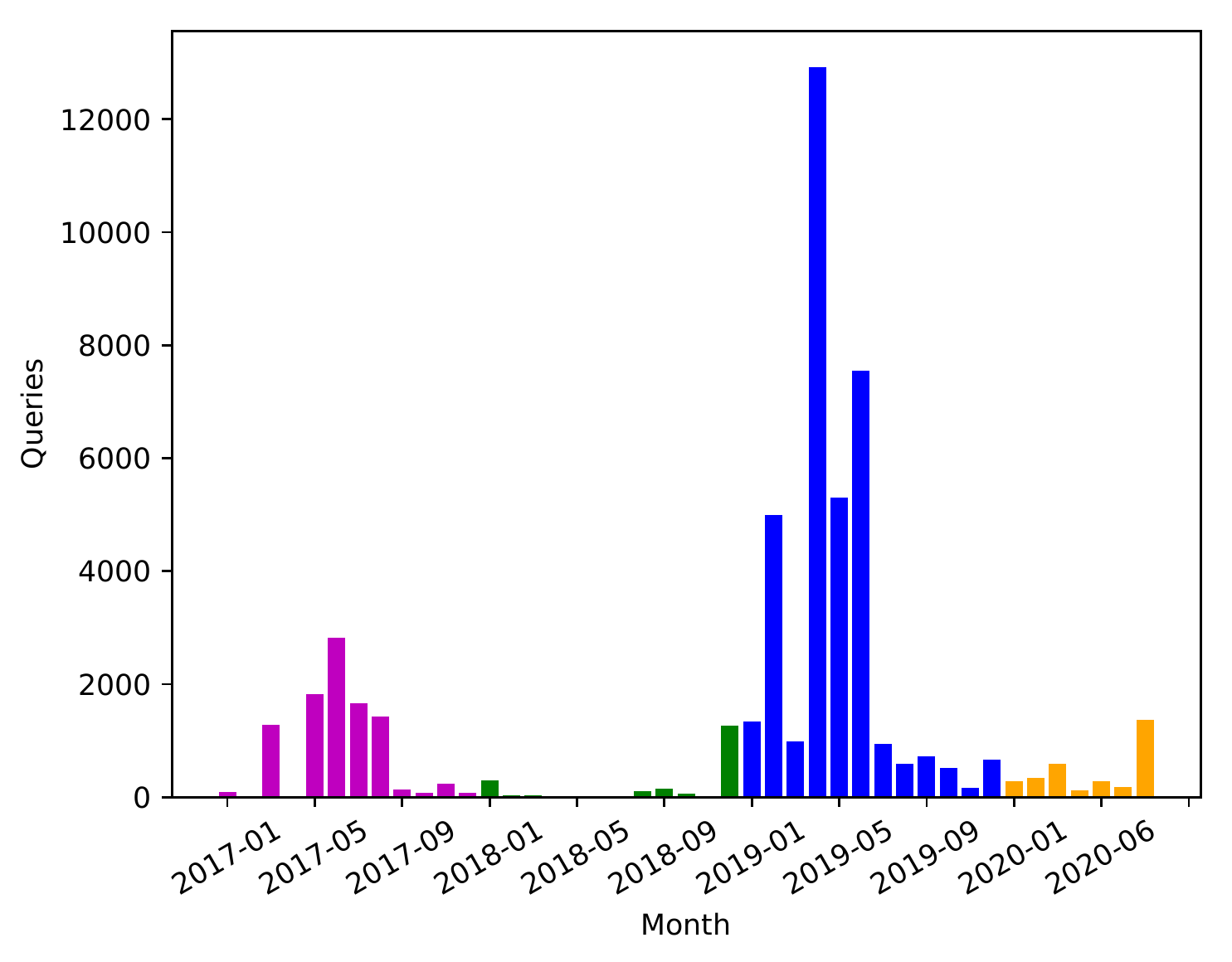}
\label{fig:mslusage}
}
\caption{Number of monthly queries for HiRISENet and MSLNet
  classifications over 3.5 years (colors distinguish years).}
\end{figure*}

HiRISENet and MSLNet generate classifications that
enable \ac{PDS} users to quickly find images of interest via
content-based search.  The public interface to the PDS image archives
is the PDS Image
Atlas\footnote{\url{https://pds-imaging.jpl.nasa.gov/search/}}. 
Users can select an instrument (e.g., HiRISE) and filter the
images to only contain a particular class of interest.  To enable this kind of
search, we applied HiRISENet and MSLNet to the
full archive of images collected by the relevant instruments on Mars.
These archives total \num{65091} HiRISE and \num{1057912} MSL images, far
more than the labeled subsets used for training and evaluation.  
\comment{from Steven: we probably want to double check the HiRISE and MSL counts.}
Figure~\ref{fig:atlas} shows the Atlas user view of a HiRISE image
with all confidently classified craters enclosed in red bounding
boxes.  Craters that are small, faint, degraded, or distorted are less
likely to pass the confidence threshold, but those identified as
craters are highly reliable.
In response to user requests, we added the ability to download a
file that contains the latitude and longitude coordinates of each
detected landmark, using
PDSC\footnote{\url{https://github.com/JPLMLIA/pdsc}} to convert from
pixel to geographic coordinates.

Classifying all HiRISE images took about five days on a GPU system and
yielded \num{29608} landmarks 
with a posterior probability of at least $0.9$,
from classes that were not ``Other''.
We also filtered out predictions for ``Spider'' or
``Swiss cheese'' at latitudes outside of the south polar
region~\cite{aye:polar19}.  This total
represents an 81\% increase over the number of classified landmarks
available in the first classifier release~\cite{wagstaff:deepmars18}.
%
MSLNet classified \num{136967}
images with a posterior probability of at least $0.9$. 
Both classifiers are integrated into the data ingestion pipeline for
the Atlas.  As new data is delivered from HiRISE or MSL, the images
are automatically processed and tagged with their predicted
classes. 

\comment{
Running our HiRISENet Model over the 159,623 images on the Atlas took
approximately 5 days. Only images classified with greater than or equal to 90
percent confidence are labeled. In addition, we employ a filter for Spiders and
Swiss cheese that are at improbable latitudes (outside the South Polar
region).
}

We track the number of Atlas queries that make use of HiRISENet and
MSLNet classifications.  As shown in Figure~\ref{fig:mrousage},
HiRISENet exhibits regular and increasing usage over time.  
The most popular HiRISE class to be queried is ``Crater''.
MSLNet shows more
varied activity (Figure~\ref{fig:mslusage}), dominated by heavy usage
in early 2019 when the number of queries for ``Wheel''
increased by several orders of magnitude (note the difference in $y$
axis range).  
Given the small
separation in query timestamps, most likely it was the result of a large
number of scripted queries to the Atlas, which provides a public API.
It is possible that this classifier output is serving to help train
other investigators' models.

We also analyzed the top 20 domain names from which the queries came.
From January to July of 2020, we found that 40\% of these queries came
from hosts through an ISP, including Spectrum and Comcast
as well as ISPs in the U.K., the Netherlands, Romania, and Taiwan.
Another 33\% of these queries came from JPL domains, which is not
surprising given the relevance of the content to JPL projects.  A
minority of the top 20 domain queries came from the Remote Sensing
Technology Center of Japan (2\%), the University of Wyoming (1\%), and
SoftBank (Japan) (1\%), while 23\% of hostnames did not resolve to a domain.
%

Finally, we used Google Analytics to examine the global distribution of
visitors who made classification queries.  Between July 2017 (the
oldest data available) and August 2020, there were \num{62613}
visitors. The top ten countries and share of visitors were: United
States (51\%), India (6\%), United Kingdom (4\%), Germany (3\%),
Canada (2\%), France (2\%), Italy (2\%), Spain (2\%), Australia (2\%),
and Russia (2\%).  In all, visitors came from 180 different
countries.

{\bf Lessons learned.} 
The deployment of Mars image classifiers at scale has yielded several
lessons learned.  First, it is worth highlighting the challenge of defining
meaningful and relevant classes up front.  Unlike a fixed benchmark
data set, new Mars images are continually collected and new classes
could arise at any time.  Collaboration with domain experts is vital
for ensuring the relevance of the class definitions.

\begin{figure}
\centering
\includegraphics[height=2.5in]{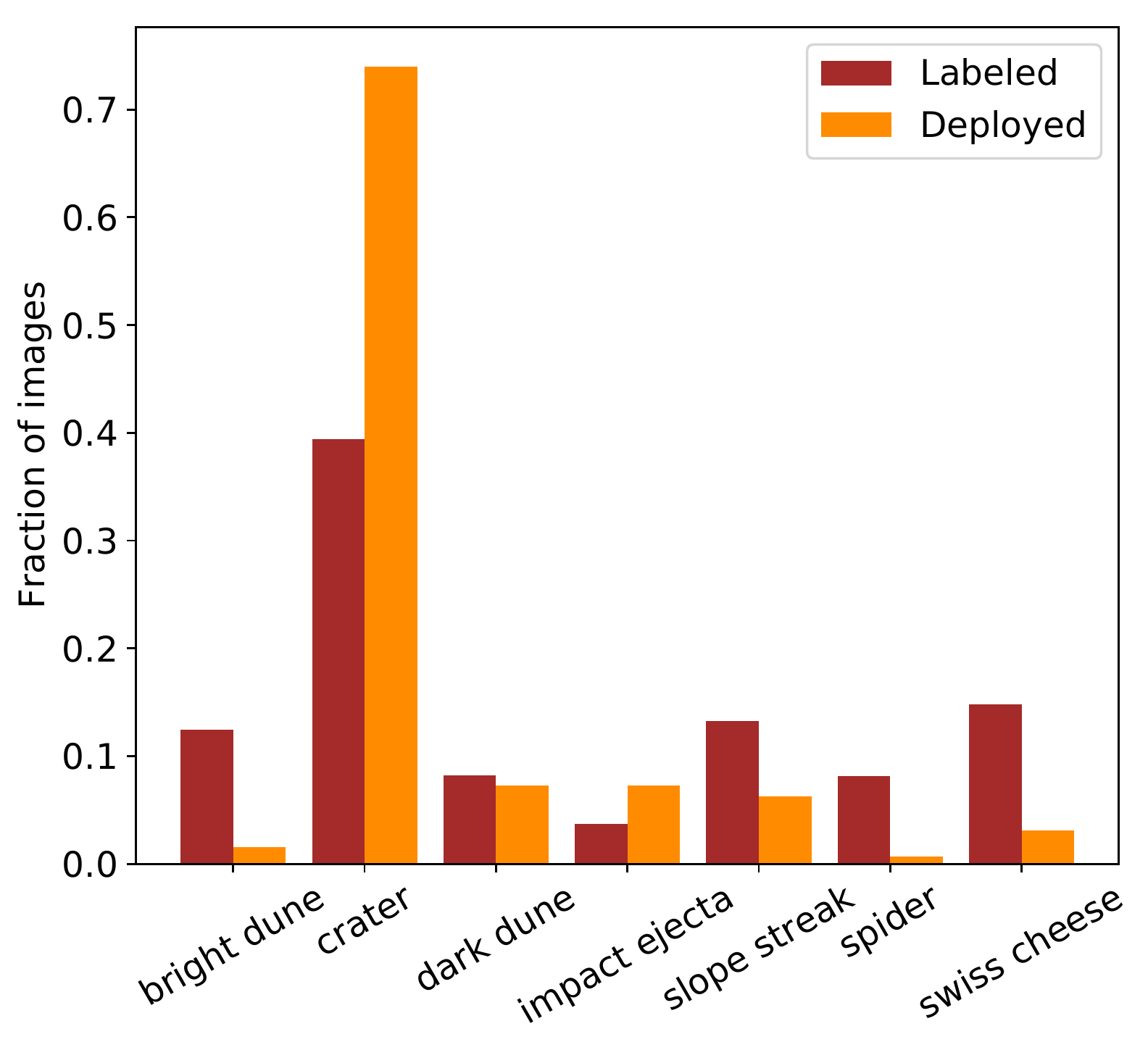}
\caption{Class distribution for HiRISE labeled data set versus
  predictions on the full archive.}
\label{fig:class_dist}
\end{figure}

Second, domain shift is an important factor in both data sets.
Figure~\ref{fig:class_dist} compares the class distribution (excluding
``Other'') for the labeled HiRISE data set (brown) to the predictions
made across the full HiRISE archive (orange).  While the ``Crater'' class is
dominant in both, its share of the images classified nearly doubles
when deployed.  There are likely two factors involved: our labeled
data set is not fully representative of all of Mars, and ``Crater''
predictions may in general be more confident and thus more likely to
pass the 0.9 threshold and be retained here.  Similar effects are seen
in the MSL data set.  We are currently investigating the use of label
shift adaptation increase the robustness of both
classifiers. 

Finally, we found that our initial simplifying assumption of one class
per image is sometimes inadequate (a crater might contain a dark slope
streak; an MSL image might contain rover parts and the horizon).
Even with guidance on class priorities, human labelers sometimes found
it difficult to select a single class label.  We plan to adopt a
multi-label approach in future versions to allow more flexibility and
reduce label noise. 

%% file: pc21-conclusion-future-work.tex
\section{Conclusions and Future Work}
\comment{
\begin{itemize}
\item Label shift correction
\item Adapt active learning methods
\end{itemize}}

This paper presents the latest updates to the deployment of machine
learning image classifiers in support of planetary science.  Two
classifiers, for orbital and surface images of Mars, have been
operating since late 2016 to enable the first content-based search
of large NASA image archives.  Usage has increased over time, and
feedback from users as well as internal assessments have guided recent
improvements.  These include acquiring additional training data to
improve minority class performance for the HiRISE classifier,
defining new classes of scientific interest for the MSL classifier,
employing calibration to increase classifier reliability, and making
landmark coordinates downloadable.
To increase our understanding of how and why users employ the machine
learning classifications of Mars images, we are investigating
minimally intrusive ways to learn more about user motivations and use
cases.
Meanwhile, the performance and scope of these classifiers continues to grow.  
Each time new images are delivered by the instruments at Mars, they
are automatically classified and added to the archive.

We are currently developing a new classifier, MERNet, that will operate on 
images collected by the two  
\ac{Pancam} instruments on the Opportunity and Spirit \ac{MER} rovers.
Based on our lessons learned, we are employing a
multi-label approach 
so multiple labels can be assigned to a single image. 
MERNet will classify all 
Pancam images using 
classes
of both science and engineering interest that were identified in a
survey conducted by the MER Data Catalog
project~\cite{cole:mdcsurvey20}.  MERNet will provide users with the
first content-based search capability for \ac{MER} images.
\comment{
The MERNet classifier was trained and evaluated using a
data set that consists of \num{68602} images spanning twenty-four classes
of both science and engineering interest that were identified in a
survey conducted by the MER Data Catalog project. 
\todo{cite the survey paper}
Eighteen classes (e.g., ``float rock'', ``layered rock'', and
``spherules'') are considered science classes, and the remaining six
classes describe rover hardware parts. 
The MERNet classifier was trained and evaluated using four \ac{CNN}
architectures: (1) transform the multi-class problem into twenty-four
one-verses-all binary problems;
(2) employ classifier chain method to allow the classifier to learn
dependencies between classes; \todo{cite the classifier chain
  paper} 
(3) investigate appending a hand-engineered feature, site location, to
the \ac{CNN} features before the classification step; 
(4) incorporate interpretability to explain the classifier's reasoning 
process \todo{cite ProtoPNet paper}.
Our preliminary evaluation results have shown that the first
architecture performs the best on the majority classes and the second
approach performs the best on minority classes.  Once the deployment
details are finalized, the classification results of MERNet will be
available on the PDS Image Atlas.
}

Finally, we plan to incorporate label shift
adaptation~\cite{alexandari:bcts20} into future 
upgrades of the Mars image classifiers.  It is evident that the data
collected by Mars instruments is not i.i.d.; class frequencies change
as spacecraft study different locations on Mars, globally from orbit
or locally via rover traverse.  By
enabling the classifiers to adapt to class probability changes,
we expect to obtain more reliable classifications.
